\documentclass{article}

\usepackage{PRIMEarxiv}

\usepackage[utf8]{inputenc} 
\usepackage[T1]{fontenc}    
\usepackage{hyperref}       
\usepackage{url}            
\usepackage{booktabs}       
\usepackage{amsfonts}       
\usepackage{nicefrac}       
\usepackage{microtype}      
\usepackage{lipsum}
\usepackage{fancyhdr}       
\usepackage{graphicx}       
\usepackage{subfigure}
\usepackage{subcaption}
\usepackage{caption}

\graphicspath{{media/}}     
\usepackage{natbib}
\title{Replicating Human Social Perception in Generative AI: Evaluating the Valence-Dominance Model}

\author{
  \textbf{Necdet Gürkan} \\
  College of Business \\
  University of Missouri - St. Louis \\
  St. Louis, Missouri \\
  \texttt{necdetgurkan@umsl.edu} \\
   \And
  Kimathi Njoki \\
 College of Business \\
  University of Missouri - St. Louis \\
  St. Louis, Missouri \\
  \texttt{kfnchx@umsl.edu} \\
 \AND
  Jordan W. Suchow \\
 School of Business \\
  Stevens Institute of Technology \\
  Hoboken, New Jersey \\
  \texttt{jws@stevens.edu} \\
}

\begin{document}
\maketitle

\begin{abstract}
As artificial intelligence (AI) continues to advance—particularly in generative models—an open question is whether these systems can replicate foundational models of human social perception. A well-established framework in social cognition suggests that social judgments are organized along two primary dimensions: valence (e.g., trustworthiness, warmth) and dominance (e.g., power, assertiveness). This study examines whether multimodal generative AI systems can reproduce this valence-dominance structure when evaluating facial images and how their representations align with those observed across world regions. Through principal component analysis (PCA), we found that the extracted dimensions closely mirrored the theoretical structure of valence and dominance, with trait loadings aligning with established definitions. However, many world regions and generative AI models also exhibited a third component, the nature and significance of which warrant further investigation. These findings demonstrate that multimodal generative AI systems can replicate key aspects of human social perception, raising important questions about their implications for AI-driven decision-making and human-AI interactions.

\end{abstract}


\section{Introduction}
The field of artificial intelligence (AI) has revolutionized many aspects of modern life, transforming both scientific research and industrial practices. In recent years, generative AI—a subset of AI that uses computational techniques to generate novel outputs by learning patterns and structures from training data—has emerged as a groundbreaking tool \citep{goodfellow2020generative}. These systems excel in tasks requiring human-like intelligence, from creative endeavors \citep{boussioux2024crowdless, epstein2023art} to complex decision-making \citep{chuma2023generative, kim2024mdagents, suri2024large}. Beyond mimicking human capabilities in specific domains, generative AI systems are now being explored for their potential to replicate and extend foundational psychological theories, offering new pathways for understanding human cognition and AI’s role in shaping social perception \citep{jiang2024evaluating, shiffrin2023probing, strachan2024testing, sartori2023language}.

At the core of generative AI lies the deep neural network architecture, which has led to a significant debate regarding its ability to replicate human cognition \citep{bowers2023deep, cichy2019deep}. These debates focus on diverse neural networks pretrained on tasks such as image classification \citep{jha2023extracting, nadler2023divergences} and natural language processing, including sentiment analysis, and machine translation \citep{tuckute2024language}. Researchers have combined deep neural network representations with both linear and cognitive models to refine these representations, effectively fine-tuning them to capture aspects of human psychological processes \citep{bhatia2019distributed, sucholutsky2024alignment, peterson2022deep, gurkan2022cultural, gurkan2023harnessing}.

Generative AI fundamentally differs from traditional AI systems in its ability to generalize beyond its training data, raising critical questions about whether human-like cognitive and psychological representations can emerge naturally, without task-specific fine-tuning. Large language models (LLMs), such as ChatGPT, Gemini, and Claude, exemplify this potential. These models, trained on billions of parameters to predict the next token in a sequence, excel in diverse tasks including reasoning \citep{huang2022towards}, text comprehension \citep{jiang2024look}, and content generation \citep{spangher2024llms}. Unlike conventional language models, their extraordinary capacity for generalization across varied contexts has sparked a growing interest in their ability to inherently encode human-like conceptual and psychological frameworks \citep{huang2023humanity, du2024human}. Thus, researchers have begun using LLMs as proxies for human participants to test widely recognized psychological theories, such as Moral Foundations Theory \citep{scherrer2024evaluating}, theory of mind \citep{strachan2024testing}, cognitive biases \citep{hagendorff2023human}, and decision-making frameworks \citep{herrmann2025standards}.

Recently, multimodal generative AI models have emerged, integrating multiple data modalities—such as text, images, audio, and video—into a single system. Examples include OpenAI’s DALL-E and GPT-4, Google DeepMind’s Gemini, and Meta’s ImageBind, demonstrating the potential of these systems to synthesize complex outputs by combining insights across different forms of data. This integration enables tasks such as generating descriptive captions for images \citep{nguyen2024improving}, creating visualizations from textual prompts \citep{yang2023reco}, and even producing video content from audio narratives \citep{sung2023sound}. Multimodal generative AI represents a significant step towards systems that more closely resemble human cognitive abilities, as humans inherently process and combine multiple sensory modalities when reasoning, communicating, or creating \citep{schulze2025visual}.

Multimodal generative models have demonstrated remarkable capabilities in replicating aspects of human cognition, including the representation of object concepts \citep{du2024human}, abstract visual reasoning \citep{camposampiero2023abstract}, and performance on perceptual psychophysical tasks \citep{marjieh2024large}. These models have also been shown to embed human-like biases. For instance, CLIP \cite{radford2021learning} reflects the values of American racial hierarchies, mirroring both implicit and explicit beliefs present in human cognition \citep{wolfe2022evidence}, and exhibits social stereotypes \cite{bianchi2023easily}. Furthermore, the trait inferences made by these models from facial images strongly correlate with human impression biases \citep{wolfe2024dataset}. Despite these advancements, the question of whether multimodal generative models can replicate broader human psychological theories remains unexplored.

In this paper, we examine whether multimodal generative AI models can replicate the valence-dominance theory of impressions of faces. The valence-dominance model \citep{oosterhof2008functional} has been fundamental in psychological research on social perception, explaining how individuals make rapid trait inferences based on facial features. By evaluating whether multimodal AI models encode these dimensions in a way that aligns with human judgments, we aim to assess their potential for modeling complex social perception processes. We tested three state-of-the-art multimodal generative AI models—Claude 3.5 Sonnet, OpenAI GPT-4 Turbo, and Gemini 1.5 Pro—by analyzing their trait inferences from facial images using principal component analysis (PCA). We then compared the extracted dimensions to those observed in human judgments across world regions, leveraging findings from the Psychological Science Accelerator project \citep{jones2021world}. This cross-cultural comparison allows us to assess whether generative AI models not only replicate the valence-dominance structure but also align with regional variations in human social perception. Our findings contribute to the broader discussion on AI’s ability to approximate human cognitive structures and shed light on its implications for understanding biases in machine-generated representations.
\section{Methods}
\subsection{Face stimuli}
The face stimuli used in this study were sourced from the Chicago Face Database \citep{ma2015chicago}, which was also utilized by the cross-cultural Psychological Science Accelerator project \citep{jones2021world} we reference to compare the alignment of multimodal generative AI models with different world regions. The dataset consists of facial images of 60 men and 60 women.
\subsection{Multimodal Generative AI}
Although some companies have attempted to discontinue features that explicitly assess individual attributes, such functionalities remain accessible. For instance, while OpenAI has stated that it no longer enables personal judgments from facial images, third-party applications like FacialAnalyzer remain publicly available.

In this study, we utilized three multimodal generative AI models accessed via API: Claude 3.5 Sonnet by Anthropic \citep{claude}, GPT-4 Turbo by OpenAI \citep{openai}, and Gemini 1.5 Pro by Google \citep{Google2024GeminiPro}. These models are widely used for a variety of tasks across both individual and organizational contexts. To evaluate their alignment with human social perception, we replicated the experiments conducted by \cite{jones2021world}, treating these three generative AI models as participants. The models were prompted with the same experimental instructions as in \cite{jones2021world} and provided ratings on a 7-point scale for each face across 13 traits.

\section{Results}
\subsection{Principal Component Analysis}
We conducted a Principal Component Analysis (PCA) following the approach of \cite{oosterhof2008functional} and \cite{jones2021world}, using orthogonal components without rotation and retaining components with eigenvalues greater than 1. First, we examined whether the first and second components of ratings generated by AI showed the same primary pattern as reported by \cite{oosterhof2008functional}. Second, we compared the number of components extracted from each generative AI model to those found in \cite{oosterhof2008functional} and across different world regions in \cite{jones2021world}. Lastly, we assessed the similarity of the first and second components using a congruence coefficient relative to different world regions.

The primary pattern reported by \cite{oosterhof2008functional} consists of a first component that correlates strongly with rated trustworthiness but not with rated dominance, and a second component that correlates strongly with rated dominance but not with rated trustworthiness. We applied the same replication criteria as \cite{jones2021world}, requiring that the first component exhibit a loading greater than 0.7 with trustworthiness and less than 0.5 with dominance, while the second component must show a loading greater than 0.7 with dominance and less than 0.5 with trustworthiness. This pattern was replicated by each generative AI model, as shown in Table 1.

We extracted the same number of components (three) for each generative AI model as almost every world region, except Africa (two) and \cite{oosterhof2008functional} (two) (Fig. 1). As noted in \cite{jones2021world}, the third component may not represent a stable or interpretable dimension, as its composition varied across AI models and explained only a small proportion of additional variance. In line with findings from world regions, traits with the highest loadings on this component also tended to crossload on the first component, suggesting that it may reflect residual variance rather than a distinct, meaningful factor (\cite{jones2021world}).

Although most world regions and generative AI models exhibited a third component, we focus our analysis on the first two components to maintain consistency with the valence-dominance model proposed by \cite{oosterhof2008functional}. This approach aligns with prior work by \cite{jones2021world}, which found that the third component varied across regions and was not clearly interpretable. By restricting our analysis to the first two components, we ensure comparability with previous studies and maintain theoretical coherence in evaluating the replication of valence and dominance in generative AI models.

\begin{table}[!h]
    \centering
    \renewcommand{\arraystretch}{1.2}
    \setlength{\tabcolsep}{10pt}
    \begin{tabular}{lcc|cc|c}
        \toprule
        \multicolumn{1}{l}{\textbf{Generative AI Model}} & \multicolumn{2}{c}{\textbf{Component 1}} & \multicolumn{2}{c}{\textbf{Component 2}} & \textbf{Replicated} \\
        \cmidrule(lr){2-3} \cmidrule(lr){4-5}
        & \textbf{Trustworthy} & \textbf{Dominant} & \textbf{Dominant} & \textbf{Trustworthy} & \\
        \midrule
        Oosterhof and Todorov & 0.94 & -0.24 & 0.92 & -0.06 \\
        Gemini 1.5 Pro & 0.81 & 0.39 & 0.85 & -0.34 & Yes \\
        Claude 3.5 Sonnet & 0.87 & 0.22 & 0.87 & -0.18 & Yes \\
        OpenAI GPT-4 Turbo & 0.81 & 0.12 & 0.86 & 0.08 & Yes \\
        \bottomrule
    \end{tabular}
    \caption{Replication of Oosterhof and Todorov’s valence-dominance model in a generative AI model was determined based on component loadings: the first component had to exhibit a loading greater than 0.7 with trustworthiness and less than 0.5 with dominance, while the second component had to show a loading greater than 0.7 with dominance and less than 0.5 with trustworthiness (\cite{jones2021world}).}
    \label{tab:replication_criteria}
\end{table}

As shown in Table 2, our results indicate that the first component was consistent across world regions and multimodal generative AI models, whereas the second component exhibited variability. This discrepancy may be influenced by the presence of the third component, which could be absorbing variance differently across models and regions. Future research should further investigate the nature and significance of this third component to determine whether it captures meaningful aspects of social perception or represents residual variance.

\begin{figure}[!h] 
    \centering
    \includegraphics[width=1.0\textwidth]{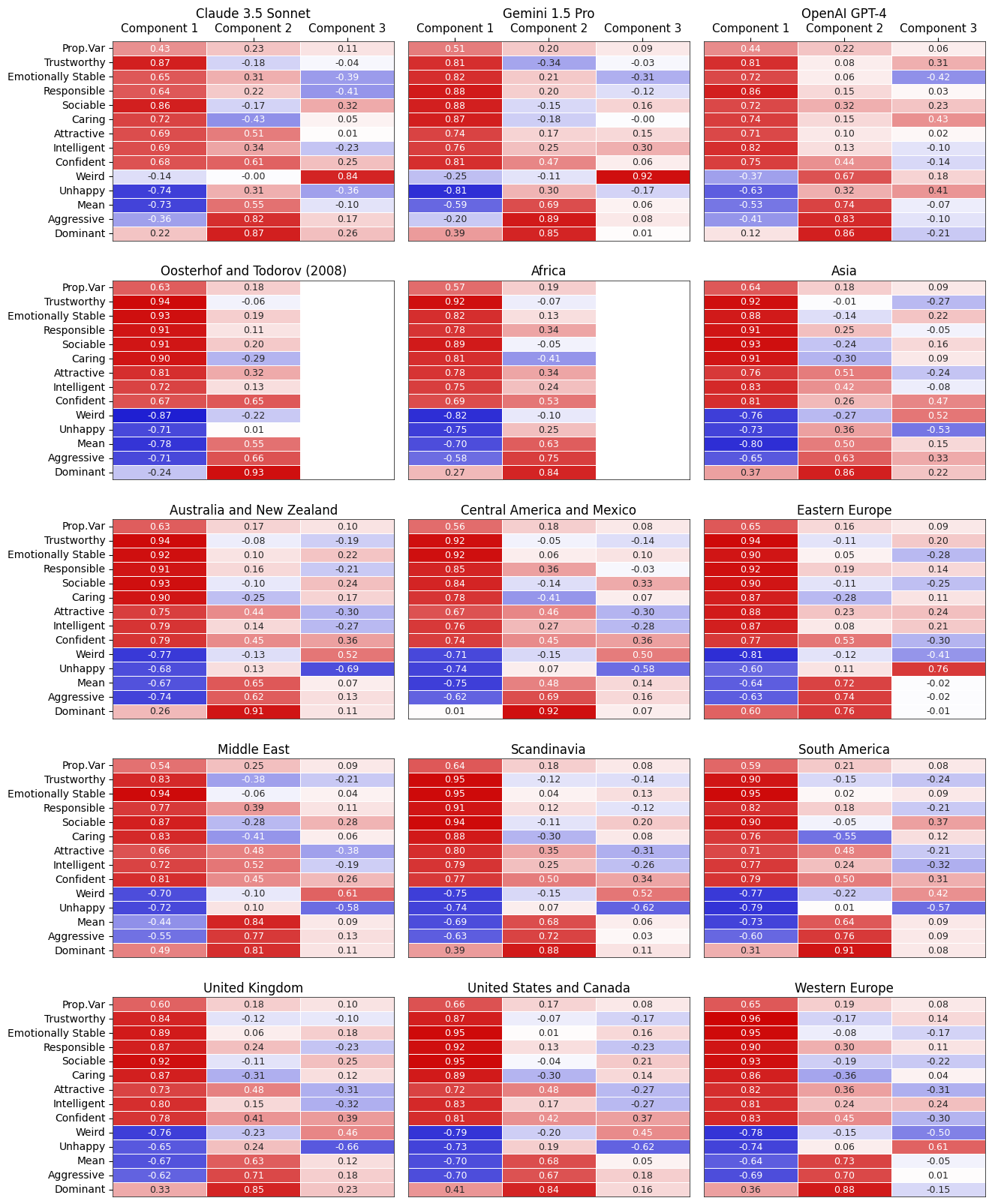} 
    \caption{PCA loading matrices for each world region and generative AI model. Positive loadings are shaded in red, while negative loadings are shaded in blue, with darker shades indicating stronger loadings. The proportion of variance explained by each component (Prop. Var) is displayed at the top of each table.}
    \label{fig:example}
\end{figure}

\begin{table}[h]
    \centering
    \small 
    \renewcommand{\arraystretch}{1.0}
    \caption{Tucker's coefficient of congruence for different regions across two principal components in three different multimodal Generative AI models.}
    \label{tab:tucker_congruence_combined}
    
    \begin{minipage}{\linewidth}
        \centering
        \captionof*{table}{(a) OpenAI GPT-4 Turbo}
        \begin{tabular}{lcccl}
            \toprule
            \textbf{Region} & \multicolumn{2}{c}{\textbf{Component 1}} & \multicolumn{2}{c}{\textbf{Component 2}} \\
            \cmidrule(lr){2-3} \cmidrule(lr){4-5}
            & \textbf{Loading} & \textbf{Congruence} & \textbf{Loading} & \textbf{Congruence} \\
            \midrule
            Africa & 0.98 & Equal & 0.78 & Not similar \\
            Asia & 0.98 & Equal & 0.65 & Not similar \\
            Australia and New Zealand & 0.98 & Equal & 0.77 & Not similar \\
            Central America and Mexico & 0.98 & Equal & 0.70 & Not similar \\
            Eastern Europe & 0.97 & Equal & 0.79 & Not similar \\
            Middle East & 0.97 & Equal & 0.69 & Not similar \\
            Scandinavia & 0.98 & Equal & 0.77 & Not similar \\
            South America & 0.98 & Equal & 0.69 & Not similar \\
            United Kingdom & 0.98 & Equal & 0.73 & Not similar \\
            United States and Canada & 0.98 & Equal & 0.76 & Not similar \\
            Western Europe & 0.98 & Equal & 0.73 & Not similar \\
            \bottomrule
        \end{tabular}
    \end{minipage}
    
    \vspace{0.4cm}
    
    \begin{minipage}{\linewidth}
        \centering
        \captionof*{table}{(b) Gemini 1.5 Pro}
        \begin{tabular}{lcccl}
            \toprule
            \textbf{Region} & \multicolumn{2}{c}{\textbf{Component 1}} & \multicolumn{2}{c}{\textbf{Component 2}} \\
            \cmidrule(lr){2-3} \cmidrule(lr){4-5}
            & \textbf{Loading} & \textbf{Congruence} & \textbf{Loading} & \textbf{Congruence} \\
            \midrule
            Africa & 0.98 & Equal & 0.78 & Not similar \\
            Asia & 0.98 & Equal & 0.65 & Not similar \\
            Australia and New Zealand & 0.98 & Equal & 0.77 & Not similar \\
            Central America and Mexico & 0.98 & Equal & 0.71 & Not similar \\
            Eastern Europe & 0.97 & Equal & 0.79 & Not similar \\
            Middle East & 0.97 & Equal & 0.69 & Not similar \\
            Scandinavia & 0.98 & Equal & 0.77 & Not similar \\
            South America & 0.98 & Equal & 0.69 & Not similar \\
            United Kingdom & 0.98 & Equal & 0.73 & Not similar \\
            United States and Canada & 0.98 & Equal & 0.75 & Not similar \\
            Western Europe & 0.98 & Equal & 0.73 & Not similar \\
            \bottomrule
        \end{tabular}
    \end{minipage}
    
    \vspace{0.4cm}
    
    \begin{minipage}{\linewidth}
        \centering
        \captionof*{table}{(c) Claude 3.5 Sonnet}
        \begin{tabular}{lcccl}
            \toprule
            \textbf{Region} & \multicolumn{2}{c}{\textbf{Component 1}} & \multicolumn{2}{c}{\textbf{Component 2}} \\
            \cmidrule(lr){2-3} \cmidrule(lr){4-5}
            & \textbf{Loading} & \textbf{Congruence} & \textbf{Loading} & \textbf{Congruence} \\
            \midrule
            Africa & 0.98 & Equal & 0.78 & Not similar \\
            Asia & 0.98 & Equal & 0.65 & Not similar \\
            Australia and New Zealand & 0.98 & Equal & 0.77 & Not similar \\
            Central America and Mexico & 0.98 & Equal & 0.71 & Not similar \\
            Eastern Europe & 0.97 & Equal & 0.79 & Not similar \\
            Middle East & 0.97 & Equal & 0.69 & Not similar \\
            Scandinavia & 0.98 & Equal & 0.77 & Not similar \\
            South America & 0.98 & Equal & 0.69 & Not similar \\
            United Kingdom & 0.98 & Equal & 0.73 & Not similar \\
            United States and Canada & 0.98 & Equal & 0.75 & Not similar \\
            Western Europe & 0.98 & Equal & 0.73 & Not similar \\
            \bottomrule
        \end{tabular}
    \end{minipage}
\end{table}

\section{Discussion}
Our findings demonstrate that multimodal generative AI models can replicate key aspects of the valence-dominance structure of social perception, aligning with human trait evaluations of faces. Using PCA, we found that the first two extracted components in generative AI models closely mirrored the theoretical dimensions of valence (trustworthiness) and dominance, as originally proposed by \cite{oosterhof2008functional} and validated across world regions by \cite{jones2021world}. This suggests that generative AI models encode foundational patterns of social perception similarly to human participants, reinforcing their potential as computational proxies for studying human cognition.

The close correspondence between AI-extracted trait dimensions and human judgments across world regions supports the idea that generative AI models internalize and reproduce cognitive frameworks fundamental to social perception. However, while the first two components exhibited high similarity, we also observed a third component in generative AI models, mirroring the additional component identified in most world regions by \cite{jones2021world}. The presence of this third component raises important questions about whether AI models capture subtle variations in facial trait inferences that are present in human judgments but were not emphasized in the original two-dimensional model.

The ability of generative AI models to replicate valence and dominance structures has significant implications for AI-driven decision-making, particularly in applications involving facial perception, such as automated hiring, security screening, and social robotics. If AI systems encode and utilize trait inferences similar to humans, they may reinforce or even amplify existing biases in social evaluations. Understanding how these models process and categorize facial information is therefore essential to ensuring fairness, transparency, and accountability in AI applications.

Despite strong alignment between AI and human trait inferences, several limitations warrant further investigation. First, the third component extracted in AI models and most world regions remains difficult to interpret, suggesting that additional dimensions of social perception may be at play. Future research should explore whether this third component reflects culturally specific aspects of facial evaluation or artifacts of PCA decomposition.

\bibliography{references}

\end{document}